\documentclass[
hf,
]{ceurart}

\usepackage{listings}
\usepackage[nolist]{acronym}
\usepackage{tikz}
\usepackage{pgfplots}
\usepackage{csquotes}
\usepackage{cleveref}
\usepackage{subcaption}

\usetikzlibrary{arrows.meta, backgrounds}

\usepgfplotslibrary{patchplots}
\usepgfplotslibrary{fillbetween}
\pgfplotsset{%
    layers/standard/.define layer set={%
        background,axis background,axis grid,axis ticks,axis lines,axis tick labels,pre main,main,axis descriptions,axis foreground%
    }{
        grid style={/pgfplots/on layer=axis grid},%
        tick style={/pgfplots/on layer=axis ticks},%
        axis line style={/pgfplots/on layer=axis lines},%
        label style={/pgfplots/on layer=axis descriptions},%
        legend style={/pgfplots/on layer=axis descriptions},%
        title style={/pgfplots/on layer=axis descriptions},%
        colorbar style={/pgfplots/on layer=axis descriptions},%
        ticklabel style={/pgfplots/on layer=axis tick labels},%
        axis background@ style={/pgfplots/on layer=axis background},%
        3d box foreground style={/pgfplots/on layer=axis foreground},%
    },
    compat = 1.3,
}

\begin{document}

\begin{acronym}
	\acro{SAT}[SAT]{Boolean satisfiability}
	\acro{MaxSAT}[MaxSAT]{Maximum Satisfiability}
	\acro{PWMS}[PWMS]{partial, weighted \ac{MaxSAT}}
	\acro{CNF}[CNF]{conjunctive normal form}
	\acro{IPASIR}[IPASIR]{re-entrant incremental satisfiability application program interface}
	\acro{IPAMIR}[IPAMIR]{re-entrant incremental MaxSAT solver application program interface}
	\acro{XOR}[XOR]{exclusive-OR}
	\acro{MIP}[MIP]{mixed integer programming}
	\acro{MSE}[MSE]{MaxSAT evaluation}
	\acro{MQT}[MQT]{Munich Quantum Toolkit}
	\acro{QCC}[QCC]{quantum color code}
	\acro{RHS}[RHS]{right-hand side}
	\acro{CDCL}[CDCL]{conflict-driven clause learning}
	\acro{SMT}[SMT]{satisfiability modulo theories}
	\acro{LOP}[LOP]{Lights Out Puzzle}
	\acro{UP}[UP]{unit propagation}
\end{acronym}
%

\copyrightyear{2024}
\copyrightclause{Copyright for this paper by its authors.
	Use permitted under Creative Commons License Attribution 4.0
	International (CC BY 4.0).}

\conference{Presented at the \href{https://www.pragmaticsofsat.org/2024/}{15th International Workshop on Pragmatics of SAT (PoS 2024)}}

\title{IGMaxHS -- An Incremental MaxSAT Solver with Support for XOR Clauses} 

\author[1]{Ole Lübke}[orcid=0000-0001-5962-6583, email=ole.luebke@tuhh.de]

\address[1]{Hamburg University of Technology (TUHH), Institute for Software Systems, Germany}

\begin{abstract}
	Recently, a novel, MaxSAT-based method for error correction in quantum computing has been proposed that requires both
	incremental MaxSAT solving capabilities and support for XOR constraints, but no dedicated MaxSAT solver fulfilling
	these criteria existed yet.
	We alleviate that and introduce IGMaxHS, which is based on the existing solvers iMaxHS and GaussMaxHS,
	but poses fewer restrictions on the XOR constraints than GaussMaxHS.
	IGMaxHS is fuzz tested with \texttt{xwcnfuzz}, an extension of \texttt{wcnfuzz}
	that can directly output XOR constraints.
	As a result, IGMaxHS is the only solver that reported neither incorrect unsatisfiability verdicts nor invalid
	models nor incoherent cost-model combinations in a final fuzz testing comparison of all three solvers with
	10000 instances.
	We detail the steps required for implementing Gaussian elimination on XOR constraints in CDCL SAT solvers,
	and extend the recently-proposed \acl{IPAMIR} to allow for incremental addition of XOR constraints.
	Finally, we show that IGMaxHS is capable of decoding quantum color codes through simulation with the Munich
	Quantum Toolkit.
\end{abstract}


\maketitle

\section{Introduction}\label{sec:intro}

Modern, incrementally-usable \ac{SAT} solvers have enabled us to solve its even more complex extensions, such as
\ac{MaxSAT} \cite{davies2011,morgado2013},
as well as practical problems in different application domains, e.g., bounded model checking and hardware validation
\cite{nadel2014a,whittemore2001}, more efficiently.
The \ac{IPASIR} \cite{balyo2016a} describes the common functionality of such solvers and has become the de-facto
standard interface for \ac{SAT}-based applications.
Recently, as an optimization counterpart to \ac{IPASIR}, the \ac{IPAMIR} has been introduced to enable solving
increasingly complex constrained optimization problems \cite{niskanen2022}.
Usually, (incremental) (Max)SAT solvers expect their input in \ac{CNF}, which can result in prohibitively large formulas
for applications where \ac{XOR} is an important operation, e.g., cryptography \cite{soos2009} or hashing-based
approximate model counting \cite{soos2021}.
A single \ac{XOR} constraint of length $n$ results in $2^{n-1}$ constraints of length $n$ in \ac{CNF}.
However, a set of \ac{XOR} constraints forms a set of linear equations that can be solved by Gaussian elimination,
which can be integrated in (Max)SAT solvers to resolve \ac{XOR} constraints more quickly \cite{soos2021,soos2009}.

Off-the-shelf (Max)SAT solvers are well-known for their efficiency,
and employed in novel applications where either no dedicated algorithm exists yet,
or where they can even outperform existing solutions \cite{yoshihiro2016,zha2023}.
A recent such example is decoding \acp{QCC} for error correction in quantum computing \cite{berent2023}.
However, this \ac{MaxSAT}-based method requires both incrementality \emph{and} support for \ac{XOR} constraints,
but no off-the-shelf dedicated \ac{MaxSAT} solver currently provides both.
The Z3 \ac{SMT} solver does provide both of these features, yet usually dedicated MaxSAT solvers are more efficient
on MaxSAT instances than general optimization solvers \cite{ansotegui2017}.

We aim to alleviate that and introduce \emph{IGMaxHS} (IncrementalGaussMaxHS)\footnote{source code available from
	\url{https://collaborating.tuhh.de/cda7728/incremental-gaussmaxhs}},
a novel, incremental \ac{MaxSAT} solver with support for incrementally adding \ac{XOR} constraints.
The solver is based on a combination of \emph{iMaxHS} \cite{niskanen2022} and \emph{GaussMaxHS} \cite{soos2021},
which are themselves both based on (different versions of) \emph{MaxHS} \cite{bacchus2017,bacchus2021a}.
Yet, IGMaxHS is more than the sum of its parts:
\begin{itemize}
	\item We lift two restrictions imposed by GaussMaxHS:
	      \begin{itemize}
		      \item GaussMaxHS rejects \ac{XOR} constraints of length shorter than three at any point in time in the
		            solving process, i.e., in the input formula, but also when a constraint shrinks after \ac{UP}.
		            IGMaxHS handles such constraints by adding a unit or two binary \ac{CNF} constraints to the
		            underlying \ac{SAT} solver when encountering an \ac{XOR} constraint of length one or two respectively.
		      \item A distinctive feature of MaxHS is how it identifies constraints that can efficiently be
		            resolved by an integrated \ac{MIP} solver \cite{davies2013}.
		            However, this can violate correctness when \ac{XOR} constraints are present.
		            Hence, it is disabled in GaussMaxHS.
		            IGMaxHS does support this feature and ensures correctness by validating solutions obtained
		            from the \ac{MIP} solver with the integrated \ac{SAT} solver.
		            If the solution was invalid, it learns a new constraint from the conflict reported by the \ac{SAT}
		            solver.
	      \end{itemize}
	\item In our final round of fuzz testing with 10000 randomly-generated instances,
	      IGMaxHS never reported wrong solutions, while MaxHS, iMaxHS,
	      and GaussMaxHS may report unsatisfiability for satisfiable instances,
	      or output incorrect models.
\end{itemize}

In the next section, we introduce the terminology and concepts that the subsequent parts of the paper build on.
\Cref{sec:combine} describes the development process of IGMaxHS.
We detail the steps required for implementing Gaussian elimination on XOR constraints in \ac{CDCL} \ac{SAT}
solvers and \ac{MaxSAT} solvers, and introduce a generic C++ template header interface that captures the functionalities
a \ac{SAT} solver must provide for such an implementation in \cref{sec:gaussinterface}.
To enable fuzz testing IGMaxHS, we extend \texttt{wcnfuzz} \cite{paxian2023},
which can already generate \ac{XOR} constraints, but encodes them in \ac{CNF}.
The extension is described in \cref{sec:fuzz} alongside the employed testing strategy and test results.
\Cref{sec:ixor} introduces an extension to the \ac{IPAMIR} to support incremental addition of \ac{XOR} constraints in a
way that maintains compatibility with existing implementations.
We include IGMaxHS in the \ac{MQT} \cite{berent2023} and show that it is indeed capable of decoding \acp{QCC} through
simulation in \cref{sec:qcc}.
Finally, we summarize the paper and give an outlook on future work in \cref{sec:summary}.

\section{Preliminaries}

When referring to \emph{variables}, we mean Boolean variables that can take the values $true$ or $false$
(or $1$ or $0$, respectively).
A \emph{literal} is either a variable $v$, or the negation of a variable, denoted by $\neg v$.
A \emph{clause} is a disjunction of literals, the \emph{length of a clause} is its number of literals, e.g.,
$(a \lor \neg b \lor c)$ is a clause of length three.
Clauses of length one are also called \emph {unit clauses} or just \emph{units}.
An \emph{\ac{XOR} clause} is an \ac{XOR} chain of literals, e.g.,
$(a \oplus \neg b \oplus c)$ is an \ac{XOR} clause of length three.
\emph{\Acf{SAT}} is the problem of deciding whether for a given propositional logic formula there exists an assignment
such that the formula evaluates to true.
Such an assignment is called a \emph{satisfying assignment} or \emph{model}.
A propositional logic formula is in \emph{\acf{CNF}} if it is a conjunction of clauses, e.g.,
$a \land (\neg b \lor c)$ is in \ac{CNF}.
In this paper \emph{\acf{MaxSAT}} means \ac{PWMS}.
Given a set of \emph{hard clauses}, and a set of \emph{soft clauses}, each associated with a \emph{weight},
\ac{PWMS} is the problem of finding a model that satisfies all hard clauses, and maximizes the weight of satisfied
soft clauses.

\emph{\Acf{UP}} is the process of removing a unit from a set of clauses in the following way \cite{davis1960}:
remove the literal from each clause in which it is falsified, and remove each clause in which it is satisfied.
It is a core procedure in modern \emph{\acf{CDCL}} \ac{SAT} solvers,
which commonly proceed as follows \cite{marques-silva2021}:
Initially, \ac{UP} is applied until there are no more units to propagate
(if there are conflicting units, unsatisfiability can already be determined).
Then, as long as not all variables are assigned and unsatisfiability has not yet been determined,
a new \emph{decision level} is introduced, a variable is selected and assigned a value, and \ac{UP} is applied.
If \ac{UP} finds a conflict, it is analyzed to find out which decision caused it,
and to learn a so-called \emph{conflict clause},
which enables non-chronological \emph{backtracking} to the conflict-causing decision \cite{marques-silva1995}.
When conflict analysis finds that the conflict originates at decision level zero, unsatisfiability is determined.
Many \ac{SAT} solvers can execute this algorithm under \emph{assumptions}, i.e., a set of literals,
which are propagated during the initial applications of \ac{UP} to fix them to the user-provided truth values.

\Ac{SAT} solving under assumptions is vital to many \ac{MaxSAT} solving algorithms \cite{davies2011,morgado2013};
here we focus on the hitting set approach implemented in MaxHS
\cite{bacchus2017,bacchus2021a,bacchus2022a,davies2011}.
A \emph{core} is a subset of the soft clauses of the \ac{MaxSAT} instance that, in conjunction with the hard
clauses, is unsatisfiable.
Cores can be extracted from \ac{SAT} solvers using the above-mentioned assumption mechanism.
Given a set of cores, a \emph{hitting set} is a set of soft clauses that contains at least one clause from each of the
cores.
A \emph{minimum cost hitting set} is a hitting set where the sum of the weights of its clauses, i.e., its cost,
is smaller than for any other hitting set.
The hitting set approach for \ac{MaxSAT} is based on the fact that, given a minimum cost hitting set and a model
that satisfies the instance, except for the clauses in the hitting set, the minimum cost of the instance is equal to
the cost of the hitting set.
The iterative algorithm maintains a set of cores $C$, which is initially empty.
In each iteration, it calculates a minimum cost hitting set $H$ for $C$, and checks whether the instance,
excluding the clauses from $H$, is satisfiable.
If so, it can terminate.
Otherwise, the \ac{SAT} solver has found a new core that is added to $C$ before the next iteration.
Minimum cost hitting sets can be computed effectively with integer linear programming or \acf{MIP}
\cite{davies2011, davies2013}.

\section{Combining iMaxHS and GaussMaxHS\label{sec:combine}}


Because IGMaxHS combines features of iMaxHS and GaussMaxHS,
both of which are based on MaxHS, a logical first step is to combine both solvers on source code level.
Yet, the underlying MaxHS versions differ, and so do the integrated \ac{SAT} solvers.
IGMaxHS is based on MaxHS 4 \cite{bacchus2021a} and uses CaDiCaL \cite{biere2020},
while GaussMaxHS is based on MaxHS 3 \cite{bacchus2017} and uses MiniSat \cite{sorensson2010}.
Furthermore, MaxHS (and iMaxHS) still use some of the data structures of MiniSat,
most notably the \texttt{Lit struct} for representing literals.
With so many complex and different components, the combination process is necessarily error-prone and demands
thorough testing.

As a first step, we copied all files that only occur in either one of the two solver code bases,
and files that are shared by both, but have binary equal content.
Next, we turned to the files shared by both programs, but which had different content.
We used a standard \texttt{diff} tool to merge them manually.
During this phase we re-enabled the \texttt{seedtype} command line argument of MaxHS (disabled in GaussMaxHS), which,
in essence, determines to what extent the built-in \ac{MIP} solver can be used to solve the \ac{MaxSAT} instance
\cite{davies2013}.
We also encountered an assertion that would let GaussMaxHS crash when it encounters \ac{XOR} constraints shorter
than three literals, which we removed.

\Ac{XOR} constraints behave similar to \ac{CNF} constraints during \ac{UP} \cite{soos2009}:
falsified literals are removed (and an empty \ac{XOR} clause signals unsatisfiability),
but satisfied literals do not immediately satisfy the entire clause.
Instead, they invert the (implicitly-maintained) \ac{RHS} of the clause.
The \ac{RHS} of an XOR-clause determines whether it should evaluate to true or false.
As an example, consider the \ac{XOR} clause $a \oplus b \oplus c$.
If $a$ is falsified, $b \oplus c$ must be true ($a$ is removed).
If $a$ is satisfied, $b \oplus c$ must be false (the \ac{RHS} is inverted), which is equivalent to $\neg b \oplus c$,
or $b \oplus \neg c$ must be true (allowing for implicitly maintaining the \ac{RHS} by inverting a literal).
Lifting the restriction on the length of \ac{XOR} clauses is especially important when such clauses can be added
incrementally, because some literals may already have fixed values, resulting in shorter \ac{XOR} constraints.
\ac{XOR} constraints of length one are exactly the same as \ac{CNF} constraints of the same length, and thus units that
can immediately be propagated.
For \ac{XOR} clauses of length two the cost of converting them to \ac{CNF} is low, so for any \ac{XOR} clause
$a \oplus b$ we add the equivalent \ac{CNF} clauses $(a \lor b) \land (\neg a \lor \neg b)$ when the \ac{RHS} is true,
and $(a \lor \neg b) \land (\neg a \lor b)$ when the \ac{RHS} is false.

When we integrated the Gaussian elimination code of GaussMaxHS into iMaxHS we found that it often interfaces with
MiniSat, which is not present anymore in iMaxHS, so the calls need to be translated to CaDiCaL.
To facilitate that process, and to get an overview of which information from the \ac{SAT} solver is required during the
Gaussian elimination procedure, we devised a generic interface as a C++ template header and replaced all \ac{SAT}
solver calls with calls to this interface.
The header is described in \cref{sec:gaussinterface}.
As a next step%
, to ensure correctness,
we fuzz tested the solver obtained from combining iMaxHS and GaussMaxHS.
We report more details on the approach and our findings in \cref{sec:fuzz}.
After the fuzz testing and debug loop, the solver fulfills the \ac{IPAMIR} specification and can resolve \ac{XOR}
constraints.
Still, \ac{XOR} constraints cannot be added incrementally yet (cf. \cref{sec:ixor}).

\subsection{Generic Interface for Integrating Gaussian Elimination\label{sec:gaussinterface}}

As mentioned above, the functions to perform Gaussian elimination on the \ac{XOR} clauses query the \ac{SAT} solver
for information and also trigger certain \ac{SAT} solving routines.
We have captured all functionality the \ac{SAT} solver needs to provide in a generic C++ header file, shown in
\Cref{lst:gausscompat} in \cref{sec:code}.
All 39 functions are designed as function templates and many use \texttt{auto}matic return type deduction to maximize
flexibility and allow for integration with different solvers.
Not all of them are strictly required for implementing Gaussian elimination; some are included for convenience
(they can be expressed in terms of the other functions), and some are helpful for debugging.
In the following we give an overview to highlight which functionalities a \ac{SAT} solver needs to provide for Gaussian
elimination to be integrated.

One set of functions (cf. ll. 12 -- 89) is concerned with obtaining information from the \ac{SAT} solver:
the number of variables in the formula, whether the formula has already been determined to be (un)satisfiable,
the current decision level, the trail of decisions (and its length), the current values of variables (and literals),
on which decision level a variable has been assigned, and the number of \ac{XOR} clauses.
Furthermore, functions to convert between variables, literals, and corresponding zero-based indices are required.

The \ac{XOR} clauses are actually managed by the \ac{SAT} solver (in this architecture).
This is also true for the Gauss matrices, yet the \ac{SAT} solver needs to provide read and write access to them and the
\ac{XOR} clauses alike (cf. ll. 91 -- 99).
The interface also expects the \ac{SAT} solver to be able to \enquote{clean} the \ac{XOR} clauses (cf. ll. 101 -- 105),
i.e., to remove literals which are assigned as described above (\cref{sec:combine}).

Finally, the Gaussian elimination code needs a way to tell the solver that it has determined the formula to be
unsatisfiable, trigger \ac{UP}, enqueue literals for \ac{UP}, backtrack to arbitrary decision levels,
and add new (\ac{CNF} conflict) clauses to the solver (cf. ll. 107 -- 130).

\subsection{Fuzz Testing with \texttt{xwcnfuzz}\label{sec:fuzz}}

To ensure that IGMaxHS does not output incorrect results, we employed fuzz testing, which can effectively uncover faults
in \ac{MaxSAT} solvers \cite{paxian2023}.
Our testing routine is built upon the existing fuzz tool \texttt{wcnfuzz} \cite{paxian2023},
which can output \ac{MaxSAT} instances in the WCNF format used in the
\aclp{MSE}\footnote{\url{https://maxsat-evaluations.github.io/2023/rules.html\#input}} (\acsp{MSE})\acused{MSE}.
Yet, to find problems related to \ac{XOR} clauses, we need the tool to output such,
which it was originally not capable of (\ac{XOR} clauses are not part of the WCNF format).
However, \texttt{wcnfuzz} actually generates \ac{XOR} clauses of length three and four, but encodes them in \ac{CNF}.
Therefore, we extended \texttt{wcnfuzz} and added a new command line switch \texttt{-{}-xor},
which instructs the tool to output hard \ac{XOR} clauses like, e.g., $(v_1 \oplus v_2 \oplus v_3)$ as
\texttt{x h 1 2 3 0}.
There is no direct support for soft \ac{XOR} clauses, yet those are realized using a hard \ac{XOR} clause and a soft
unit clause with an activation literal \cite{soos2021}.
For a soft \ac{XOR} clause like, e.g., $(v_1 \oplus v_2 \oplus v_2)$ with weight $10$, an activation literal,
e.g., $v_4$ is added to form a hard clause \texttt{x h 4 1 2 3 0} and a soft clause \texttt{10 -4 0}.
We refer to the resulting tool as \texttt{xwcnfuzz}, and to the resulting output file format as XWCNF.

We performed multiple testing rounds, starting with only 10 instances, and multiplying the number of instances with 10
when no incorrect results were found, until IGMaxHS was able to reliably solve 10000 instances without producing
erroneous results%
.
For comparison, we used MaxHS (in the \ac{MSE} 2021 version, which iMaxHS is based on), iMaxHS and GaussMaxHS,
which were executed on equivalent input instances in their respective input formats
(e.g., the pre-MSE-2022 (X)WCNF input
format\footnote{\url{https://maxsat-evaluations.github.io/2021/rules.html\#input}}).
Additionally, we relied on the
\texttt{verify\_soln}\footnote{\url{https://bitbucket.org/fbacchus/maxsat_benchmarks_code_base/src/master/}} tool from
the \acp{MSE} to verify the output from the solvers.
A timeout was set to 5 seconds per instance for each solver.
In the following we report the errors and remedies we found during testing, and finally compare all four solvers in
a final test run.

One class of errors is related to the fact that the \ac{MIP} solver in IGMaxHS is not aware of the \ac{XOR} clauses
(remember that GaussMaxHS restricted the use of the \ac{MIP} solver to prevent such problems).
There is a Boolean flag \texttt{allClausesSeeded} that signals whether the \ac{MIP} solver has complete knowledge of
the instance at hand, which must never be \texttt{true} in presence of \ac{XOR} clauses.
Solutions from the \ac{MIP} solver are used to update the lower and upper bounds the MaxHS algorithm maintains.
To prevent IGMaxHS from using invalid models for these updates,
we employ the \ac{SAT} solver to verify the model and reject the update if it does not satisfy the hard (\ac{XOR})
clauses.
In such cases, the \ac{SAT} solver yields a conflict clause that is then added to the \ac{MIP} solver to prevent the
conflict from occurring again.

The remaining errors concern the integration of Gaussian elimination in the \ac{SAT} solver.
GaussMaxHS integrates Gaussian elimination in the \ac{CDCL} loop of MiniSat, when \ac{UP} did not find any
conflicts.
We followed this approach, but found that \ac{UP} is used for many different purposes in iMaxHS and CaDiCaL.
One example is the function to perform \ac{UP} under user-provided assumptions (\texttt{find\_up\_implicants}),
which has been added to CaDiCaL in iMaxHS.
For this to be correct, it is required that \ac{UP} is also performed on the \ac{XOR} clauses.
Consequently, we integrated Gaussian elimination directly in the \texttt{propagate} function of CaDiCaL.
This, however, requires the Gauss matrices to be initialized on any execution path that may trigger \ac{UP}.
Yet, this design gave rise to another fault: when a clause is added to the \ac{SAT} solver, the solver may reduce
it to a unit (or it may be a unit clause originally), which triggers \ac{UP}, which now triggers Gaussian
elimination, from which the solver may learn a conflict - yet the memory to add new, original clauses and conflict
clauses is shared in CaDiCaL, and we need to ensure not to overwrite the original clause.
We also found that it can be necessary to reinitialize the Gauss matrices when backtracking to decision level zero,
but only if the solver is not currently analyzing a conflict.
Otherwise, constraints that were resolved by now-invalid variable assignments are not taken into account anymore.
Because the \ac{SAT} solver may find some variables to be obsolete, it may renumber the variables, and that renumbering
also needs to be applied to the \ac{XOR} clauses, and the corresponding matrices need to be reinitialized.
Finally, the tests triggered a situation where conflict analysis on the \ac{XOR} clauses yields a unit clause that
is falsified under the current model.
The original code in GaussMaxHS expects conflicts to be at least of length two and would crash in that scenario.
Instead, IGMaxHS instructs the \ac{SAT} solver to backtrack to the decision level before the unit was assigned,
and then enqueues it for \ac{UP}.


\begin{table}[b]
	\centering
	\caption{Results from final test run\label{tab:fuzzresults} with 10000 instances generated by \texttt{xwcnfuzz}}
	\enquote{correct} means the solver reported a verifiable solution.
	\enquote{wrong optimum} means the solver asserted the optimum was found, but another solver found a better solution.
	\enquote{wrong unsat} means the solver reported unsatisfiability for a satisfiable instance.
	\enquote{not verified} means the reported solution either does not satisfy the hard clauses, or the reported cost
	does not match the actual cost of the model.
	\enquote{crash} means the solver exited without any verdict before the timeout.
	\enquote{timeout} means the solver execution was forcibly terminated after 5 sec.

	\begin{tabular}{l|rrrrrr}\hline
		solver     & correct       & wrong optimum & wrong unsat & not verified & crash      & timeout    \\ \hline
		IGMaxHS    & 9365          & \textbf{0}    & \textbf{0}  & \textbf{0}   & 6          & 638        \\
		GaussMaxHS & 5080          & \textbf{0}    & 1           & 4            & 4701       & 214        \\
		iMaxHS     & \textbf{9827} & \textbf{0}    & 167         & \textbf{0}   & \textbf{4} & \textbf{2} \\
		MaxHS      & 9825          & \textbf{0}    & 167         & 1            & 5          & \textbf{2} \\ \hline
	\end{tabular}
\end{table}

After the above-mentioned errors were rectified, we executed a final round of fuzz testing with 10000 instances
and a timeout of 5 seconds per instance for each solver.
The results are shown in
\Cref{tab:fuzzresults}.
GaussMaxHS was able to solve 50.8\% of the instances correctly, and crashed on 47.0\%.
This is due to the assertion on the length of \ac{XOR} clauses discussed previously.
Note, however, that all \ac{XOR} clauses in the input instances had length three or four, and so the removal of
assigned literals during the solving procedure must have caused these crashes.
iMaxHS and MaxHS, being very similar in their implementations, expectedly produced almost the same results,
and solved 98.3\% correctly.
iMaxHS has a slight advantage, because it was able to solve two more instances.
However, both incorrectly reported unsatisfiability of the given instance in 167 cases, usually for instances with
very high weights.
This error is also included in the fault classification by Paxian et al., however MaxHS did not exhibit it in their
testing \cite{paxian2023}.
They used a newer version of MaxHS (from the \ac{MSE} 2022 \cite{bacchus2022a}),
which most likely is the reason for this difference.
Their work also reports detailed results for Z3, which they found may yield wrong optimum values, wrong unsatisfiability
verdicts, and output that does not withstand the verifier, too.
Finally, IGMaxHS solved 93.7\% of the instances correctly,
and is the only solver that did not output incorrect solutions.
It crashed on 6 instances
(due to an inconsistent \ac{MIP} model which leads to the lower bound exceeding the upper bound in the hitting-set
algorithm), which is comparable to MaxHS and iMaxHS.
Yet, it did not come to any conclusion before the timeout for 638 instances, which is the highest timeout count among
all solvers.
Most likely, these timeouts can be attributed to the additional \ac{SAT} solver calls and frequent reinitialization
of the Gauss matrices.

\section{Incrementally Adding XOR Clauses\label{sec:ixor}}

With the \ac{IPAMIR}, hard \ac{CNF} clauses are added literal by literal with the function
\texttt{void ipamir\_add\_hard(void* solver, int32\_t lit\_or\_zero)}.
Using \texttt{0} for the \texttt{lit\_or\_zero} argument signals the end of the clause.
To extend \ac{IPAMIR} to allow for the incremental addition of \ac{XOR} clauses, we decided to use the same function.
For signalling whether the clause that is currently added is a \ac{CNF} or \ac{XOR} clause, we equipped the
function with an additional Boolean argument with a default value: \texttt{bool is\_xor = false}.
This way, compatibility with existing implementations is maintained, and the \ac{IPAMIR} is not cluttered with
additional functions.

Our implementation closely follows that of iMaxHS for adding hard \ac{CNF} clauses and that of GaussMaxHS for adding
hard \ac{XOR} clauses.
The \texttt{is\_xor} flag only needs to be set on the clause-finalizing call of the function, as all other calls only
store the provided literals in a \texttt{vector} (as in iMaxHS).
When the clause is finalized, we first check its size.
If it is a unit, we fall back to the clause adding routine of iMaxHS.
Otherwise, duplicate literals, and literals whose values are already fixed are removed from the clause as in GaussMaxHS.
If, afterward, the clause is empty and its \ac{RHS} is false, it is already satisfied and ignored.
Otherwise, the clause is added to the solver (if it is empty, this signals unsatisfiability).
Finally, we check whether the current model satisfies the clause.
If not, the current model is invalidated.

\section{Decoding Quantum Color Codes\label{sec:qcc}}

Recently, a \ac{MaxSAT} based method to decode \acfp{QCC} has been proposed that requires a solver that is capable
of incrementally updating the problem instance and has support for \ac{XOR} clauses \cite{berent2023}.
As no dedicated \ac{MaxSAT} solver with these features existed at the time, the authors turned to the Z3 \ac{SMT}
solver \cite{demoura2008}.
In this section we demonstrate that IGMaxHS is capable of decoding \acp{QCC} by integrating it with the \acf{MQT}
\cite{berent2023} and executing the corresponding simulations.

The \ac{QCC} decoding method is based on an analogy to the \ac{LOP} \cite{berent2023,sutner1989}.
Explaining the method in detail is beyond the scope of this paper; a brief introduction to the \ac{LOP} is sufficient
to understand the resulting \ac{MaxSAT} instances.
Consider a grid of $n \times n$ lights, each equipped with a light switch.
Toggling a switch turns off the corresponding light and all directly adjacent lights (i.e., excluding diagonals).
Given an initial configuration of turned-on lights, find a minimum-size set of switches that turn all lights off.
Note, that toggling a switch twice is equivalent to not using it at all, so each switch in the resulting set must only
be toggled once \cite{sutner1989}.

Let $L$ be the set of lights, and $S$ be the set of switches (in this simplified example $|L| = |S|$, in general
the underlying graph structure can be more complex).
Introducing a Boolean variable $v_s, s \in |S|$ for the state of each switch, a function
$\mathcal{S}: L \rightarrow S^*$
that, given a light, yields the set of switches that can toggle that light, and the initial configuration
$\mathcal{S}_{init}: L \rightarrow \{0, 1\}$, we can model the \ac{LOP} as a \ac{MaxSAT} instance with hard clauses
$\forall l \in L: \bigoplus_{s \in \mathcal{S}(l)} v_s = \mathcal{S}_{init}(l)$ and soft clauses
$\forall s \in S: \neg v_s$ with weight $1$ \cite{berent2023}.
Introducing $k = |\mathcal{S}(l)| - 1$ helper variables $h_1, \dots, h_k$ allows for rewriting the hard clauses as
\begin{align*}
	v_1 \oplus h_1 & = S_{init}(l)                                             \\
	h_i            & = v_{i+1} \oplus h_{i+1} & \text{\ for\ } i = 1, \dots, k \\
	h_k            & = v_k,
\end{align*}
such that the \ac{XOR} clauses are split into smaller clauses of which only the first must be updated with the current
configuration during error correction \cite{berent2023}.

\begin{figure}
	\begin{subfigure}{.45\textwidth}

\begin{tikzpicture}[/tikz/background rectangle/.style={fill={rgb,1:red,1.0;green,1.0;blue,1.0}, fill opacity={1.0}, draw opacity={1.0}}, show background rectangle]
\begin{axis}[point meta max={nan}, point meta min={nan}, legend cell align={left}, legend columns={1}, title={}, title style={at={{(0.5,1)}}, anchor={south}, font={{\fontsize{14 pt}{18.2 pt}\selectfont}}, color={rgb,1:red,0.0;green,0.0;blue,0.0}, draw opacity={1.0}, rotate={0.0}, align={center}}, legend style={color={rgb,1:red,0.0;green,0.0;blue,0.0}, draw opacity={1.0}, line width={1}, solid, fill={rgb,1:red,1.0;green,1.0;blue,1.0}, fill opacity={1.0}, text opacity={1.0}, font={{\fontsize{8 pt}{10.4 pt}\selectfont}}, text={rgb,1:red,0.0;green,0.0;blue,0.0}, cells={anchor={center}}, at={(0.02, 0.98)}, anchor={north west}}, axis background/.style={fill={rgb,1:red,1.0;green,1.0;blue,1.0}, opacity={1.0}}, anchor={north west}, xshift={1.0mm}, yshift={-1.0mm}, width={.9\textwidth}, height={.2\textheight}, scaled x ticks={false}, xlabel={physical error rate}, x tick style={color={rgb,1:red,0.0;green,0.0;blue,0.0}, opacity={1.0}}, x tick label style={color={rgb,1:red,0.0;green,0.0;blue,0.0}, opacity={1.0}, rotate={0}}, xlabel style={at={(ticklabel cs:0.5)}, anchor=near ticklabel, at={{(ticklabel cs:0.5)}}, anchor={near ticklabel}, font={{\fontsize{10 pt}{12 pt}\selectfont}}, color={rgb,1:red,0.0;green,0.0;blue,0.0}, draw opacity={1.0}, rotate={0.0}}, xmode={log}, log basis x={10}, xmajorgrids={true}, xmin={0.001}, xmax={0.1}, xticklabels={{$10^{-3}$,$10^{-2}$,$10^{-1}$}}, xtick={{0.001,0.01,0.1}}, xtick align={inside}, xticklabel style={font={{\fontsize{8 pt}{10.4 pt}\selectfont}}, color={rgb,1:red,0.0;green,0.0;blue,0.0}, draw opacity={1.0}, rotate={0.0}}, x grid style={color={rgb,1:red,0.0;green,0.0;blue,0.0}, draw opacity={0.1}, line width={0.5}, solid}, axis x line*={left}, x axis line style={color={rgb,1:red,0.0;green,0.0;blue,0.0}, draw opacity={1.0}, line width={1}, solid}, scaled y ticks={false}, ylabel={average solving time [ms]}, y tick style={color={rgb,1:red,0.0;green,0.0;blue,0.0}, opacity={1.0}}, y tick label style={color={rgb,1:red,0.0;green,0.0;blue,0.0}, opacity={1.0}, rotate={0}}, ylabel style={at={(ticklabel cs:0.5)}, anchor=near ticklabel, at={{(ticklabel cs:0.5)}}, anchor={near ticklabel}, font={{\fontsize{10 pt}{12 pt}\selectfont}}, color={rgb,1:red,0.0;green,0.0;blue,0.0}, draw opacity={1.0}, rotate={0.0}}, ymajorgrids={true}, ymin={0}, ymax={385.6419}, yticklabels={{$0$,$100$,$200$,$300$}}, ytick={{0.0,100.0,200.0,300.0}}, ytick align={inside}, yticklabel style={font={{\fontsize{8 pt}{10.4 pt}\selectfont}}, color={rgb,1:red,0.0;green,0.0;blue,0.0}, draw opacity={1.0}, rotate={0.0}}, y grid style={color={rgb,1:red,0.0;green,0.0;blue,0.0}, draw opacity={0.1}, line width={0.5}, solid}, axis y line*={left}, y axis line style={color={rgb,1:red,0.0;green,0.0;blue,0.0}, draw opacity={1.0}, line width={1}, solid}, colorbar={false}]
    \addplot[color={rgb,1:red,0.0;green,0.6056;blue,0.9787}, name path={bf7c9430-df1f-4418-b734-8901c8d700e6}, draw opacity={1.0}, line width={1}, solid]
        table[row sep={\\}]
        {
            \\
            0.001  0.5159  \\
            0.01  0.44439999999999996  \\
            0.1  0.9995  \\
        }
        ;
    \addlegendentry {d = 3}
    \addplot[color={rgb,1:red,0.8889;green,0.4356;blue,0.2781}, name path={80335305-069d-4234-bbac-9e4386e20b82}, draw opacity={1.0}, line width={1}, solid]
        table[row sep={\\}]
        {
            \\
            0.001  1.4863  \\
            0.01  1.7757  \\
            0.1  4.817  \\
        }
        ;
    \addlegendentry {d = 5}
    \addplot[color={rgb,1:red,0.2422;green,0.6433;blue,0.3044}, name path={567fe897-da73-4b51-86be-2bad982b842e}, draw opacity={1.0}, line width={1}, solid]
        table[row sep={\\}]
        {
            \\
            0.001  5.9985  \\
            0.01  5.3635  \\
            0.1  40.6525  \\
        }
        ;
    \addlegendentry {d = 7}
    \addplot[color={rgb,1:red,0.7644;green,0.4441;blue,0.8243}, name path={4ae9075d-8434-4096-af90-0abe06d37a0e}, draw opacity={1.0}, line width={1}, solid]
        table[row sep={\\}]
        {
            \\
            0.001  17.956599999999998  \\
            0.01  15.5815  \\
            0.1  198.9216  \\
        }
        ;
    \addlegendentry {d = 9}
    \addplot[color={rgb,1:red,0.6755;green,0.5557;blue,0.0942}, name path={3b2ffaf1-733c-4aef-b533-c12ff54757b5}, draw opacity={1.0}, line width={1}, solid]
        table[row sep={\\}]
        {
            \\
            0.001  35.6387  \\
            0.01  43.417300000000004  \\
            0.1  385.6419  \\
        }
        ;
    \addlegendentry {d = 11}
\end{axis}
\end{tikzpicture}
		\caption{Solver runtime in dependence of the error rate, for different code sizes}
	\end{subfigure}
	\begin{subfigure}{.45\textwidth}

\begin{tikzpicture}[/tikz/background rectangle/.style={fill={rgb,1:red,1.0;green,1.0;blue,1.0}, fill opacity={1.0}, draw opacity={1.0}}, show background rectangle]
\begin{axis}[point meta max={nan}, point meta min={nan}, legend cell align={left}, legend columns={1}, title={}, title style={at={{(0.5,1)}}, anchor={south}, font={{\fontsize{14 pt}{18.2 pt}\selectfont}}, color={rgb,1:red,0.0;green,0.0;blue,0.0}, draw opacity={1.0}, rotate={0.0}, align={center}}, legend style={color={rgb,1:red,0.0;green,0.0;blue,0.0}, draw opacity={1.0}, line width={1}, solid, fill={rgb,1:red,1.0;green,1.0;blue,1.0}, fill opacity={1.0}, text opacity={1.0}, font={{\fontsize{8 pt}{10.4 pt}\selectfont}}, text={rgb,1:red,0.0;green,0.0;blue,0.0}, cells={anchor={center}}, at={(0.02, 0.98)}, anchor={north west}}, axis background/.style={fill={rgb,1:red,1.0;green,1.0;blue,1.0}, opacity={1.0}}, anchor={north west}, xshift={1.0mm}, yshift={-1.0mm}, width={.9\textwidth}, height={.2\textheight}, scaled x ticks={false}, xlabel={distance}, x tick style={color={rgb,1:red,0.0;green,0.0;blue,0.0}, opacity={1.0}}, x tick label style={color={rgb,1:red,0.0;green,0.0;blue,0.0}, opacity={1.0}, rotate={0}}, xlabel style={at={(ticklabel cs:0.5)}, anchor=near ticklabel, at={{(ticklabel cs:0.5)}}, anchor={near ticklabel}, font={{\fontsize{10 pt}{12 pt}\selectfont}}, color={rgb,1:red,0.0;green,0.0;blue,0.0}, draw opacity={1.0}, rotate={0.0}}, xmajorgrids={true}, xmin={3}, xmax={11}, xticklabels={{$11$,$3$,$5$,$7$,$9$}}, xtick={{11.0,3.0,5.0,7.0,9.0}}, xtick align={inside}, xticklabel style={font={{\fontsize{8 pt}{10.4 pt}\selectfont}}, color={rgb,1:red,0.0;green,0.0;blue,0.0}, draw opacity={1.0}, rotate={0.0}}, x grid style={color={rgb,1:red,0.0;green,0.0;blue,0.0}, draw opacity={0.1}, line width={0.5}, solid}, axis x line*={left}, x axis line style={color={rgb,1:red,0.0;green,0.0;blue,0.0}, draw opacity={1.0}, line width={1}, solid}, scaled y ticks={false}, ylabel={average solving time [ms]}, y tick style={color={rgb,1:red,0.0;green,0.0;blue,0.0}, opacity={1.0}}, y tick label style={color={rgb,1:red,0.0;green,0.0;blue,0.0}, opacity={1.0}, rotate={0}}, ylabel style={at={(ticklabel cs:0.5)}, anchor=near ticklabel, at={{(ticklabel cs:0.5)}}, anchor={near ticklabel}, font={{\fontsize{10 pt}{12 pt}\selectfont}}, color={rgb,1:red,0.0;green,0.0;blue,0.0}, draw opacity={1.0}, rotate={0.0}}, ymajorgrids={true}, ymin={0}, ymax={385.6419}, yticklabels={{$0$,$100$,$200$,$300$}}, ytick={{0.0,100.0,200.0,300.0}}, ytick align={inside}, yticklabel style={font={{\fontsize{8 pt}{10.4 pt}\selectfont}}, color={rgb,1:red,0.0;green,0.0;blue,0.0}, draw opacity={1.0}, rotate={0.0}}, y grid style={color={rgb,1:red,0.0;green,0.0;blue,0.0}, draw opacity={0.1}, line width={0.5}, solid}, axis y line*={left}, y axis line style={color={rgb,1:red,0.0;green,0.0;blue,0.0}, draw opacity={1.0}, line width={1}, solid}, colorbar={false}]
    \addplot[color={rgb,1:red,0.0;green,0.6056;blue,0.9787}, name path={2f6bdabd-c24e-43bb-a2be-18c1dabca0cd}, draw opacity={1.0}, line width={1}, solid]
        table[row sep={\\}]
        {
            \\
            3.0  0.5159  \\
            5.0  1.4863  \\
            7.0  5.9985  \\
            9.0  17.956599999999998  \\
            11.0  35.6387  \\
        }
        ;
    \addlegendentry {p = 0.001}
    \addplot[color={rgb,1:red,0.8889;green,0.4356;blue,0.2781}, name path={8993c042-3e41-4f8f-9287-f18e3115e833}, draw opacity={1.0}, line width={1}, solid]
        table[row sep={\\}]
        {
            \\
            3.0  0.44439999999999996  \\
            5.0  1.7757  \\
            7.0  5.3635  \\
            9.0  15.5815  \\
            11.0  43.417300000000004  \\
        }
        ;
    \addlegendentry {p = 0.01}
    \addplot[color={rgb,1:red,0.2422;green,0.6433;blue,0.3044}, name path={e342ae6a-adfa-4636-8e3f-9f19d7d588b5}, draw opacity={1.0}, line width={1}, solid]
        table[row sep={\\}]
        {
            \\
            3.0  0.9995  \\
            5.0  4.817  \\
            7.0  40.6525  \\
            9.0  198.9216  \\
            11.0  385.6419  \\
        }
        ;
    \addlegendentry {p = 0.1}
\end{axis}
\end{tikzpicture}
		\caption{Solver runtime in dependence of the code size, for different error rates}
	\end{subfigure}
	\caption{\ac{QCC} decoding simulation results\label{fig:sim}}
\end{figure}
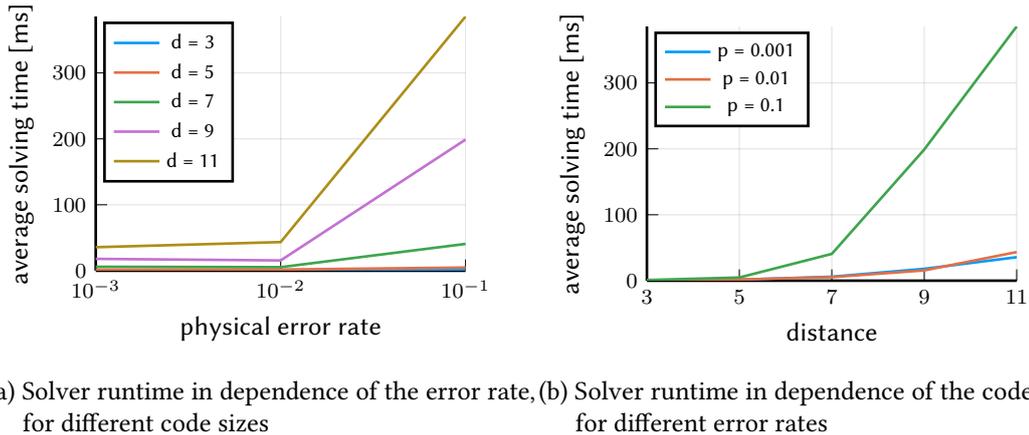

In the \ac{MQT}, the underlying graph structure is not an $n \times n$ square, but an equilateral triangle with $d$
switches on each side that is filled with hexagons which represent the lights.
Each hexagon corner is a switch that toggles the state of all adjacent hexagons.
The \ac{MQT} has built-in checks to verify the solutions obtained from the solver.
We performed simulations with $d \in \{3, 5, 7, 9, 11\}$ and physical error rates (essentially bit-flip probabilities)
$p \in \{0.001, 0.01, 0.1\}$, similar to Berent et al. \cite{berent2023}, and measured the runtime of IGMaxHS.
The result are shown in \Cref{fig:sim}.
In its current state, IGMaxHS decodes \acp{QCC} approximately 100 times slower than Z3
(ca. 4 ms on average for $d = 11$, $p = 0.1$).
The \ac{MQT} reported no decoding errors, and the increase in runtime for larger values for $d$ and $p$ is in line
with the original results obtained with Z3 \cite{berent2023}.
In conclusion, IGMaxHS is indeed capable of decoding \acp{QCC}, but currently is slower than the existing solution.

\section{Summary \& Future Work\label{sec:summary}}

IGMaxHS is a novel \ac{MaxSAT} solver that implements the \acf{IPAMIR} and additionally supports
(the incremental addition of) \ac{XOR} clauses.
Through fuzz testing with \texttt{xwcnfuzz} we ensured it neither reports incorrect unsatisfiability verdicts nor
invalid models nor incoherent cost-model combinations.
On the one hand, this distinguishes IGMaxHS from other, similar solvers, as outlined in \cref{sec:fuzz}.
On the other hand, it is often slower than these solvers.
We described its development process, detailed the steps required to implement Gaussian elimination in (Max)SAT
solvers, and captured which functionalities they need to provide in a C++ template header.
Finally, we demonstrated that IGMaxHS is practically capable of decoding \acfp{QCC}.

In the future, we aim to remove the performance penalties introduced by additional \ac{SAT} solver calls and
matrix reinitializations that are currently required for correctness.
One option could be to encode \ac{XOR}-clauses in the \ac{MIP} as well to require fewer (down to zero) validations
of models obtained from the \ac{MIP} solver, but while reducing \ac{SAT} solver runtime, this would increase \ac{MIP}
solver runtime.
Increasing the speed of IGMaxHS should enable us to run more thorough simulations of decoding \acp{QCC},
and to conduct a more in-depth comparison to Z3 with regard to decoding speed and robustness.
Finally, to increase confidence in the results provided by IGMaxHS, we would like to equip it with proof logging
capabilities.
This is standard for \ac{SAT} solvers, and several ways to generate and check unsatisfiability proofs for \ac{CNF}
extended with \ac{XOR} exist \cite{gocht2021,soos2023,tan2024},
yet proof logging has only recently been proposed for \ac{MaxSAT} solvers \cite{berg2023}.

\begin{acknowledgments}
	Many thanks to the anonymous reviewers of this paper for their constructive feedback and questions.
\end{acknowledgments}

\bibliography{IGMaxHS}

\appendix

\section{Source Code Listings\label{sec:code}}

\begin{lstlisting}[
	language=C++,
	numbers=left,
	caption={Gauss elimination compatibility interface \texttt{gauss\_compat.hpp}},
	label=lst:gausscompat]
// Copyright (c) Ole Luebke
// Licensed under MIT license
#ifndef GAUSS_COMPAT_HPP
#define GAUSS_COMPAT_HPP

#include <string>

template<typename Solver, typename Clause> class Gaussian;

namespace gauss_compat {

template<typename Solver>
inline auto get_num_variables(const Solver* solver);

template<typename Solver>
inline bool is_unsat(const Solver* solver);

template<typename Solver>
inline bool is_ok(const Solver* solver);

template<typename Solver>
inline auto get_decision_level(const Solver* solver);

template<typename Solver>
inline auto& get_trail(Solver* solver);

template<typename Solver>
inline auto get_trail_size(const Solver* solver);

template<typename Solver, typename Variable>
inline auto get_variable_value(const Solver* solver,
                               const Variable& var);

template<typename VariableValue>
inline bool is_var_unassigned(VariableValue var);

template<typename VariableValue>
inline bool is_var_assigned_true(VariableValue var);

template<typename VariableValue>
inline bool is_var_assigned_false(VariableValue var);

template<typename Solver, typename Literal>
inline auto get_literal_value(const Solver* solver,
                              const Literal& lit);

template<typename LiteralValue>
inline bool is_literal_unassigned(LiteralValue lit);

template<typename LiteralValue>
inline bool is_literal_assigned_true(LiteralValue lit);

template<typename LiteralValue>
inline bool is_literal_assigned_false(LiteralValue lit);

template<typename Literal>
inline bool does_literal_have_sign(const Literal& lit);

template<typename Literal>
inline Literal xor_sign(const Literal& lit, bool toggle);

template<typename Solver, typename Variable>
inline auto get_variable_decision_level(Solver* solver,
                                        const Variable& var);
template<typename Solver>
inline auto get_num_xor_clauses(const Solver* solver);

template<typename Variable>
inline auto variable_to_literal(const Variable& var, bool negate);

template<typename Solver, typename Literal>
inline auto literal_to_0based_varidx(const Solver* solver,
                                     const Literal& lit);

template<typename Solver, typename Literal>
inline auto literal_to_variable(const Solver* solver,
                                const Literal& lit);

template<typename Variable>
inline auto variable_to_0based_varidx(const Variable& var);

template<typename VarIdx>
inline auto _0based_varidx_to_variable(VarIdx idx);

template<typename Solver>
inline const auto& get_gauss_conf(const Solver* solver);

template<typename Solver>
inline auto get_verbosity_level(const Solver* solver);

template<typename Solver>
inline auto& get_xor_clauses(Solver* solver);

template<typename Solver>
inline auto& get_gauss_matrixes(Solver* solver);

template<typename Solver, typename Clause>
inline void new_gaussian(Solver* solver,
                         Gaussian<Solver, Clause>* gaussian);

template<typename Solver, typename Xors>
inline bool clean_xor_clauses(Solver* solver, Xors& xors);

template<typename Solver>
inline bool clean_xor_clauses(Solver* solver);

template<typename Solver>
inline void set_unsat(Solver* solver, bool unsat);

template<typename Solver>
inline bool propagate(Solver* solver);

template<typename Solver, typename Literal>
inline void enqueue(Solver* solver, const Literal& lit);

template<typename Solver, typename Literal, typename Clause>
inline void enqueue(Solver* solver, const Literal& lit, Clause* cls);

template<typename Solver, typename Level>
inline void backtrack(Solver* solver, Level level);

template<typename Solver, typename Clause>
inline auto learn_clause(Solver* solver, Clause& clause);

template<typename Solver, typename Clause>
inline void mark_clause_as_garbage(Solver* solver, Clause* clause);

template<typename Solver, typename Clause>
inline void mark_clause_as_temp_gauss(Solver* solver,
                                      Clause* clause);

template<typename TruthValue>
inline std::string truth_value_to_string(const TruthValue& val);

}

#endif
\end{lstlisting}

\end{document}